\documentclass{article}
\usepackage{spconf,amsmath,graphicx,hyperref}
\usepackage{amsfonts,amssymb}

\usepackage{booktabs}   
\usepackage[table]{xcolor} 
\usepackage{pifont}
\title{A Dual-Branch Framework for Semantic Change Detection with Boundary and Temporal Awareness}
\name{
Yun-Cheng~Li\textsuperscript{\rm 1}, 
Sen Lei\textsuperscript{\rm 1}, 
Heng-Chao~Li\textsuperscript{\rm 1}\sthanks{Corresponding author: Heng-Chao~Li. This work was supported in part by the National Natural Science Foundation of China under Grant 62271418, and in part by the Natural Science Foundation of Sichuan Province under Grants 2023NSFSC0030 and 2025ZNSFSC1154.}, 
Ke Li\textsuperscript{\rm 2} 
}
\address{
    \textsuperscript{\rm 1}School of Information Science and Technology, Southwest Jiaotong University, 611756, China.\\
    \textsuperscript{\rm 2}School of Computer Science and Technology, Xidian University, 710126, China.\\
}
\begin{document}
%
\maketitle
\begin{abstract}
Semantic Change Detection (SCD) aims to detect and categorize land-cover changes from bi-temporal remote sensing images. Existing methods often suffer from blurred boundaries and inadequate temporal modeling, limiting segmentation accuracy. To address these issues, we propose a Dual-Branch Framework for Semantic Change Detection with Boundary and Temporal Awareness, termed DBTANet. Specifically, we utilize a dual-branch Siamese encoder where a frozen SAM branch captures global semantic context and boundary priors, while a ResNet34 branch provides local spatial details, ensuring complementary feature representations. On this basis, we design a Bidirectional Temporal Awareness Module (BTAM) to aggregate multi-scale features and capture temporal dependencies in a symmetric manner. Furthermore, a Gaussian-smoothed Projection Module (GSPM) refines shallow SAM features, suppressing noise while enhancing edge information for boundary-aware constraints. Extensive experiments on two public benchmarks demonstrate that DBTANet effectively integrates global semantics, local details, temporal reasoning, and boundary awareness, achieving state-of-the-art performance. 
\end{abstract}
\section{Introduction}
\label{sec:intro}
Change detection (CD) identifies changes in objects within a region by analyzing images captured at different times. It is widely used in applications such as disaster monitoring, urban growth analysis, and environmental protection \cite{GM2015, LYC2024, like2024}. CD is divided into two categories: Binary Change Detection (BCD) and Semantic Change detection (SCD) \cite{YW2025}. At present, many studies pay attention to the BCD task, identifying the changed region through bi-temporal remote sensing images. However, BCD fails to provide detailed information on the types of changes, which limits its wide application in the field of remote sensing. In contrast, SCD can not only detect changed areas, but also recognize the specific change types, thereby constructing the ``from-to" relationships in change mapping.

In early research, statically analyzing pixel and neighborhood features are utilized to achieve SCD. The method \cite{XG2009} first detects the change of land cover type by analyzing the change vector, and then classifies the land cover of the changed areas. The kernel slow feature analysis \cite{WC2017} was utilized to fuse classification results. However, these approaches depend on accurate object segmentation, inaccurate segmentation raises the uncertainty in detecting changes \cite{Longjiang2024}.

The deep learning has significantly mitigated these issues. For example, Ding et al. \cite{DL2022} proposed a new CNN framework for SCD, in which semantic temporal features are integrated within a deep change detection unit. In \cite{DL2024}, spatio-temporal dependencies are exploited to enhance the accuracy of SCD. Jiang et al. \cite{LJ2025} proposed a novel network based on a multi-task architecture to identify potential land cover changes. Wang et al. \cite{WQ2024} proposed an SCD framework for cross-differential semantic consistency networks to mine the deep differences in instance features. Although these methods are effective, they usually regard semantic segmentation (SS) and CD as loosely connected parallel problems. Recent studies such as \cite{HLJ2024, ZJ2025} have begun to explore the interaction between SS and CD. BT-SCD \cite{TYJ2025} further introduced the Boundary Detection (BD) task as an auxiliary target to enhance the correlation between them. 

Despite these advances, two critical challenges remain: (1) blurred change boundaries, as CNN backbones tend to smooth features and lose fine-grained spatial details; and (2) insufficient temporal modeling, as bi-temporal differencing fails to capture structured dependencies, leading to misclassification of subtle or complex changes.

The Segment Anything Model (SAM) \cite{Kirillov2023} provides strong global semantic priors and boundary cues, while ResNet34 offers complementary local structural details. However, SAM alone suffers from over-smoothed representations that reduce inter-class discriminability, making it unsuitable as a standalone Siamese encoder for SCD. This motivates us to integrate SAM’s global priors with ResNet’s local precision, thereby enhancing both boundary localization and semantic discrimination.

To this end, we propose a Dual-Branch Framework for Semantic Change Detection with Boundary and Temporal Awareness (DBTANet), which incorporates three key components:
\begin{itemize}
\item We propose a novel method termed DBTANet for the identification of SCD. In contrast to prior works, DBTANet facilitates a frozen SAM branch and a light-weight ResNet34 branch to enable complementary representation learning.
\item We propose a Gaussian-smoothed Projection Module (GSPM) to refine shallow SAM features by suppressing high-frequency noise. In addition, a Bidirectional Temporal Awareness Module (BTAM) is designed to capture multi-scale change semantics and explicitly model temporal dependencies.
\item Comprehensive experiments demonstrate the robustness of DBTANet, with SeK of 65.72\% on Landsat-SCD, 24.12\% on SECOND dataset.
\end{itemize}

\begin{figure*}[!t]
  \centering
  {\includegraphics[width=\linewidth]{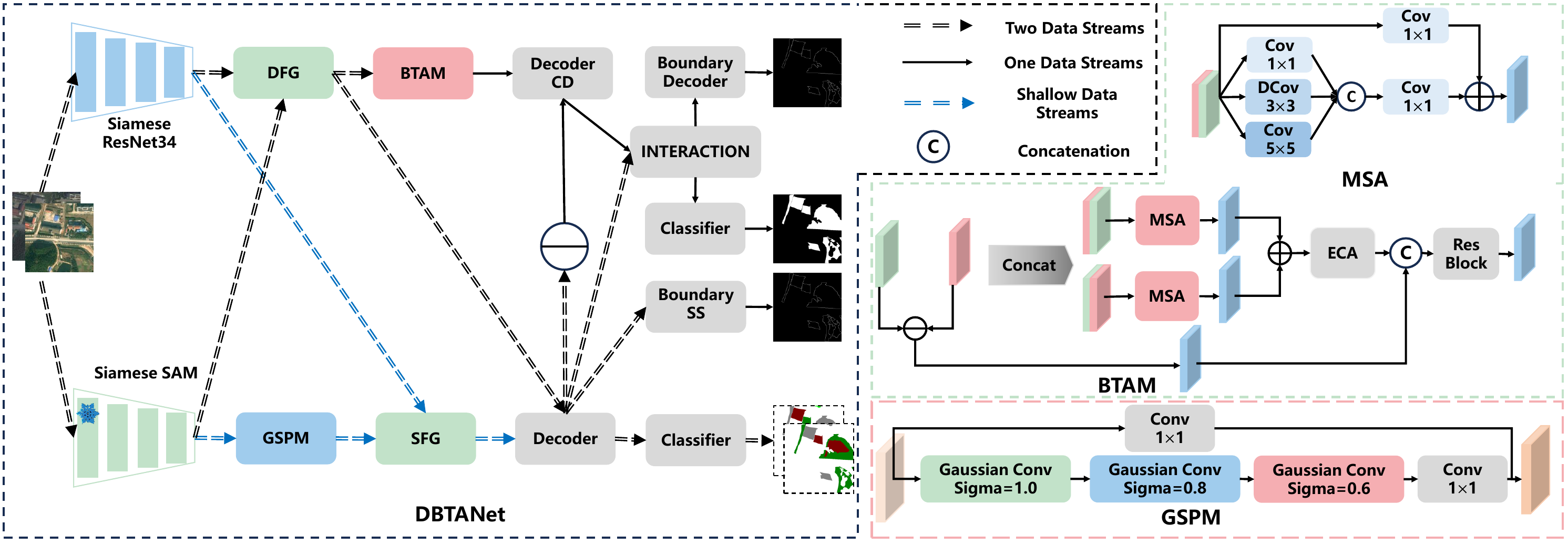}}
  \caption{The overall framework of the proposed DBTANet.}
  \label{fig:1}
\end{figure*}

\section{METHODOLOGY}
\label{sec:format}
\subsection{Overview}
The architecture of the proposed DBTANet network is illustrated in Fig. \ref{fig:1}. DBTANet is designed to address the challenges of SCD by integrating the complementary strengths of frozen SAM and ResNet34. Specifically, SAM provides global semantic priors and boundary information, while ResNet34 offers local spatial precision. A dual-branch Siamese encoder extracts features from bi-temporal images ${I_1, I_2}$, ensuring both global and local representations are captured.

The encoder of ResNet34 extracts shallow features from its shallow layer, and deep features from deep layer. Simultaneously, the frozen SAM encoder processes the same input images. SAM generates multi-scale features: shallow features from its intermediate layers, which contain rich boundary information but also high-frequency noise, and deep features from its neck layer, which includes semantic knowledge.

To effectively integrate these complementary feature streams, we introduce two key modules. The GSPM processes the noisy shallow SAM features, suppressing noise while enhancing semantically meaningful boundary structures. Additionally, feature gates enable adaptive fusion of SAM and ResNet features at shallow and deep levels, respectively. Specifically, feature gate is used to extract shallow feature map termed shallow feature gate (SFG). The Learnable feature gate is utilized to extract deep feature map called deep feature gate (DFG). They allow the model to dynamically balance fine-grained details and semantic context from SAM and ResNet34. The calculation formula is presented as follows:
\begin{equation}
    \label{eqn:1}
    \begin{aligned}
    F_{shallow} &= (1-\alpha) \times F^{res}_{shllow} + \alpha \times \operatorname{GSPM}(F^{SAM}_{shallow}),\\
    F_{deep} &= (1-\beta) \times F^{res}_{deep} + \beta \times F^{SAM}_{deep},
    \end{aligned}
\end{equation}
where $F_{shallow}$ and $F_{deep}$ represent the shallow and deep feature maps, respectively. $\alpha$ and $\beta$ indicate the shallow and deep feature gate parameters. $F^{res}_{shllow}$ and $F^{res}_{deep}$ are shallow and deep feature maps of ResNet34. $F^{SAM}_{shllow}$ and $F^{SAM}_{deep}$ are shallow and deep feature maps of SAM. 

The fused features are then processed by task-specific decoders. For SS, two separate decoders generate semantic predictions for different temporal. For CD, BTAM comprehensively models temporal dependencies and change characteristics by aggregating multi-scale features and processing them in a symmetric bidirectional manner. The CD result is then refined by a task interaction module, which utilizes the semantic feature differences to constrain the change map and computes a similarity loss to maintain consistency. The BD decoder enhances boundary precision using a Sobel-based operator, serving as an auxiliary task.

\subsection{Gaussian-smoothed Projection Module}
SAM's shallow features are rich in boundary information, but contaminated by high-frequency noise that degrades SCD performance. To address this, we design the GSPM, which progressively decreases noise and enhances boundary information. The GSPM consists of three sequential Gaussian convolution blocks with decreasing $\sigma$ (1.0, 0.8, 0.6), followed by a $1\times1$ projection and residual fusion. The larger initial $\sigma$ (1.0) effectively suppresses coarse noise, while progressively smaller $\sigma$ (0.8, 0.6) preserves increasingly finer edge details. Each block comprises a depthwise Gaussian convolution with fixed kernels followed by point-wise convolution, batch normalization, and ReLU activation. The mathematical formulation is as follows:
\begin{equation}
\label{eqn:2}
\begin{aligned}
F^{SAM}_{s} = &\operatorname{Conv}_{1\times1} ( \operatorname{GCB}_{0.6} ( \operatorname{GCB}_{0.8} ( \\
& \operatorname{GCB}_{1.0}(F^{SAM}_{shallow}) ) ) ) + \operatorname{Conv}_{1\times1}(F^{SAM}_{shallow}), \\[8pt]
\operatorname{GCB}_{\sigma}(X) = &\operatorname{Conv}_{1\times1} ( \operatorname{DepthwiseGaussianConv}_{\sigma}(X) ),
\end{aligned}
\end{equation}
where $GCB_{\sigma}(\cdot)$ denotes a Gaussian Convolution Block with parameter $\sigma$.

\subsection{Bidirectional Temporal Awareness Module}
To overcome the limitations of change detection methods that rely on simple feature differencing, we propose the BTAM. This module effectively captures change information across different scales, and models complex temporal dependencies between bi-temporal features.

The BTAM module processes the fused deep features from both temporal steps through a bidirectional architecture. First, bidirectional change representations are computed as:
\begin{equation}
\label{eqn:3}
\begin{aligned}
F^{12}_{concat} = \operatorname{MSA}(\operatorname{Concat}(F^{t1}_{deep}, F^{t2}_{deep})),\\
F^{21}_{concat} = \operatorname{MSA}(\operatorname{Concat}(F^{t2}_{deep}, F^{t1}_{deep})),
\end{aligned}
\end{equation}
where $F^{t1}_{deep}$ represents the feature map at time t1 and $\operatorname{MSA}(\cdot)$ is the Multi-Scale Aggregation block. 

The Multi-scale Aggregation (MSA) block is utilized to obtain $F^{12}_{concat}$ and $F^{21}_{concat}$ by three parallel branches:
\begin{equation}
\label{eqn:4}
\begin{aligned}
\operatorname{MSA}(X) &= \operatorname{Conv}_{1\times1} \big( \operatorname{Concat} \big( \operatorname{Conv}_{1\times1}(X), \\
& \operatorname{DConv}_{3\times3}(X), \operatorname{Conv}_{5\times5}(X) \big) \big) + \operatorname{Conv}_{1\times1}(X),
\end{aligned}
\end{equation}
where $\operatorname{DConv}_{3\times3}(\cdot)$ indicates $3\times3$ dilation convolution module while the dilation is set to 2, $\operatorname{Conv}_{5\times5}(\cdot)$ indicates $5\times5$ convolution module while padding is 2. Specifically, the $1\times1$ convolution is used for original resolution, the $5\times5$ convolution captures larger context, and $3\times3$ dilated convolution expands receptive field. This multi-scale design enables the model to capture both fine-grained and large-area changes effectively. The bidirectional features are then fused by an ECA module \cite{wang2020eca}. These features are combined with the absolute feature difference and processed through residual blocks.

This comprehensive approach allows the model to capture changes at multiple scales, and model temporal relationships symmetrically, while suppressing noise.

\begin{figure*}[!htb]
\centering
\includegraphics[width=1\linewidth]{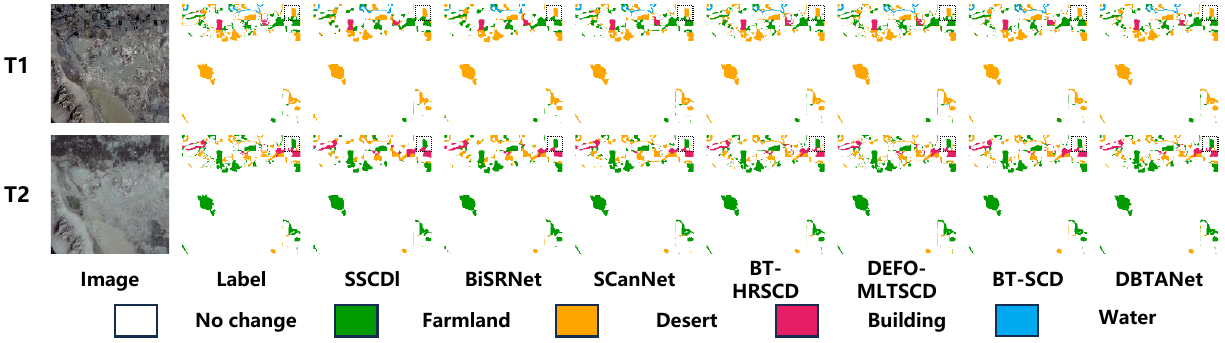}
\caption{Visual comparison on the Landsat-SCD dataset.}
\label{fig:x}
\end{figure*}

\section{EXPERIMENTS}
\label{sec:pagestyle}
\subsection{Experimental Settings}
\noindent \textbf{Datasets.}
We evaluate our method on two SCD benchmarks, \textit{i.e.,} Landsat-SCD and SECOND. The Landsat-SCD dataset contains 8,468 pairs of bi-temporal remote sensing images, each with a spatial resolution of $416 \times 416$ pixels. The SECOND dataset comprises 4,662 bi-temporal image pairs with a resolution of $512 \times 512$ pixels. Following standard practice, the Landsat-SCD dataset is partitioned into 1,455, 485, 485 image pairs for training, validation, and testing, respectively. We split SECOND dataset into 2,968 pairs for training, 847 pairs for validation, and 847 pairs for testing.

\noindent \textbf{Implementation Details}
All experiments are conducted on Ubuntu 20.04 with a single NVIDIA GeForce RTX 4090 GPU. We adopt the AdamW optimizer with an initial learning rate of 0.001 for both datasets. The batch size is set to 8. Models are trained for 80 epochs on Landsat-SCD and 50 epochs on SECOND. Standard data augmentation techniques, including random flipping and rotation, are applied. For evaluation, we employ four widely used metrics: Overall Accuracy (OA), Separated Kappa (SeK), mean Intersection-over-Union (mIoU), and F1-score for SCD (F1).   

\subsection{Experimental Results}
We compare DBTANet with six representative SCD methods, including SSCDl \cite{DL2022}, BiSRNet \cite{DL2022}, SCanNet \cite{DL2024}, BT-HRSCD \cite{FS2024}, DEFO-MLTSCD \cite{LZ2024}, and BT-SCD \cite{TYJ2025} on the Landsat-SCD and SECOND datasets, respectively.
\begin{table}[!t]
\caption{Quantitative results on Landsat-SCD dataset.}
\label{table:1}
\centering
\resizebox{\linewidth}{!}{
\begin{tabular}{l c c c c}
\toprule
Methods ~ & OA(\%) & mIoU(\%) & SeK(\%) & F1(\%) \\
\midrule
SSCDl \cite{DL2022} &94.17  &83.80  &46.22  &82.96  \\
BiSRNet \cite{DL2022} &94.34  &84.16  &47.51  &83.73 \\
SCanNet \cite{DL2024} &95.92  &88.33  &58.32  &88.19  \\
BT-HRSCD \cite{FS2024} &96.35  &89.48  &61.59  &89.38  \\
DEFO-MLTSCD \cite{LZ2024} &96.31  &89.54  &61.47  &89.18  \\
BT-SCD \cite{TYJ2025} &96.15  &88.73  &59.32  &88.57  \\
\rowcolor{blue!8} DBTANet (Ours) &\textbf{96.92}  &\textbf{90.84}  &\textbf{65.72}  &\textbf{90.90}  \\
\bottomrule
\end{tabular}}
\end{table}
\begin{table}[!t]
\centering
\caption{Quantitative results on SECOND dataset.}
\resizebox{\linewidth}{!}{
\begin{tabular}{l c c c c}
\toprule
Methods ~ & OA(\%) & mIoU(\%) & SeK(\%) & F1(\%) \\
\midrule
SSCDl \cite{DL2022} &87.40  &72.89  &22.87  &63.01  \\
BiSRNet \cite{DL2022} &87.68  &73.26  &23.59  &63.86 \\
SCanNet \cite{DL2024} &87.87  &73.28  &23.67  &63.96 \\
BT-HRSCD \cite{FS2024} &87.85  &73.34  &23.46  &63.62 \\
DEFO-MLTSCD \cite{LZ2024} &88.02  &73.37  &23.61  &63.86 \\
BT-SCD \cite{TYJ2025} &88.10  &73.45  &23.61  &63.80 \\
\rowcolor{blue!8} DBTANet (Ours) &\textbf{88.23}  &\textbf{73.74}  &\textbf{24.12}  &\textbf{64.08} \\
\bottomrule
\end{tabular}}
\label{table:2}
\end{table}
\noindent \textbf{Comparison Results on Landsat-SCD.} As shown in Table~\ref{table:1}, DBTANet consistently outperforms all competing methods across four evaluation metrics. In particular, it achieves an OA of 96.92\%, mIoU of 90.84\%, SeK of 65.72\%, and F1 of 90.90\%. Compared with the second-best method (BT-HRSCD), our framework yields substantial improvements of +4.13\% in SeK, +1.36\% in mIoU, +1.52\% in F1, and +0.57\% in OA. These results highlight that DBTANet can effectively recover fine-grained boundary details and produce semantically consistent change masks, validating the advantage of the proposed boundary-aware design.

\noindent \textbf{Comparison Results on SECOND.} Table~\ref{table:2} presents the results on the SECOND dataset. DBTANet continues to deliver competitive performance, obtaining the best mIoU (73.74\%) and SeK (24.12\%), as well as the highest F1 (64.08\%). Compared to BT-SCD, which ranks second in most metrics, our method achieves relative gains of +0.29\% in mIoU, +0.51\% in SeK, and +0.28\% in F1. These improvements, though smaller than those on Landsat-SCD, still demonstrate the robustness of DBTANet under different data distributions and domain variations.

\noindent \textbf{Visualization Results.}
As shown in Fig.~\ref{fig:x}, our DBTANet generates sharper and more accurate semantic boundaries, particularly in challenging scenes with cluttered objects or subtle changes, where conventional CNN-based backbones tend to produce blurred or incomplete masks. These visual comparisons further validate the effectiveness of incorporating SAM-guided boundary information with adaptive feature fusion and multi-scale temporal modeling.
\begin{table}[!t]
    \centering
    \caption{Effect of components in our DBTANet.}
    \label{table:3}
    \resizebox{0.9\linewidth}{!}{
        \begin{tabular}{c c c c c}
            \toprule
             \textbf{+ SAM} & \textbf{+ GSPM} & \textbf{+ BTAM} & \textbf{mIoU(\%)} & \textbf{SeK(\%)} \\ 
            \midrule
             \ding{55} & \ding{55} & \ding{55} &72.34  &22.30  \\
             \ding{51} & \ding{55} & \ding{55} &72.68  &22.76   \\
             \ding{51} & \ding{51} & \ding{55} &72.76  &23.26   \\
             \ding{51} & \ding{51} & \ding{51} &\textbf{73.74}  &\textbf{24.12}   \\
            \bottomrule
        \end{tabular}}
\end{table}

\noindent \textbf{Ablation Study}
We conduct ablation experiments on the SECOND dataset to evaluate the contributions of the SAM, GSPM, and BTAM modules (Table~\ref{table:3}). Starting from the baseline, incorporating SAM yields consistent improvements in both mIoU (from 72.34\% to 72.68\%) and SeK (from 22.30\% to 22.76\%), demonstrating the benefit of SAM features for boundary refinement. Adding the GSPM module further enhances performance, especially in boundary quality, with SeK increasing to 23.26\%. Finally, equipping the model with all three modules leads to the best results, achieving 73.74\% mIoU and 24.12\% SeK, confirming that SAM-guided features, Gaussian-smoothed projection, and multi-scale bidirectional modeling are complementary and jointly boost SCD.

\section{Conclusion}
\label{sec:majhead}
In this paper, we propose DBTANet, a semantic change detection framework that integrates the complementary strengths of SAM and ResNet34 in a dual-branch Siamese architecture. The frozen SAM provides global semantic priors and boundary information, while ResNet34 captures local spatial details. To further enhance performance, a GSPM is proposed to refine noisy shallow SAM features, and a BTAM is designed to model multi-scale change semantics and temporal dependencies. Extensive experiments on public benchmarks demonstrate that DBTANet achieves state-of-the-art performance by jointly improving global–local representation, temporal reasoning, and boundary precision.

\bibliographystyle{IEEEbib}
\bibliography{strings,refs}

\end{document}